\documentclass[letterpaper, 10 pt, conference]{ieeeconf}  % Comment this line out if you need a4paper

\IEEEoverridecommandlockouts                              % This command is only needed if 
\usepackage{graphics} % for pdf, bitmapped graphics files
\usepackage{graphicx}
\usepackage{graphicx}
\usepackage{epsfig} % for postscript graphics files
\usepackage{mathptmx} % assumes new font selection scheme installed
\usepackage{times} % assumes new font selection scheme installed
\usepackage{amsmath} % assumes amsmath package installed
\usepackage{amssymb}  % assumes amsmath package installed
\usepackage{amsmath}
\usepackage{booktabs}
\usepackage{microtype}

\usepackage{amsmath,amsfonts}
\usepackage{algorithmic}
\usepackage{algorithm}
\usepackage{float}
\title{\LARGE \bf
Dynamic Parameter Identification of a Curtain Wall Installation Robotic Arm
}

\author{Xiao Liu$^{1}$,Yunxiao Cheng$^{1}$, Weijun Wang$^{1}$, Tianlun Huang$^{1}$, Wei Feng $^{1,2*}$% <-this % stops a space
\thanks{*corresponding author. e-mail: wei.feng@siat.ac.cn}% <-this % stops a space
\thanks{$^{1}$ All authors are with Shenzhen Institute of Advanced Technology, Chinese Academy of Sciences, Shenzhen, 51800, China.
        {\tt\small  Contact: xiao.liu1@siat.ac.cn}}%
 \thanks{$^{2}$ All authors are with Shenzhen University of Advanced Technology and University of Chinese Academy of Sciences, Shenzhen, 51800, China.
        {\tt\small  Contact: wei.feng@siat.ac.cn}}%
}

\begin{document}

\maketitle
\thispagestyle{empty}
\pagestyle{empty}

%%%%%%%%%%%%%%%%%%%%%%%%%%%%%%%%%%%%%%%%%%%%%%%%%%%%%%%%%%%%%%%%%%%%%%%%%%%%%%%%
\begin{abstract}

 In the construction industry, traditional methods fail to meet the modern demands for efficiency and quality. The curtain wall installation is a critical component of construction projects. We design a hydraulically driven robotic arm for curtain wall installation and a dynamic parameter identification method. We establish a Denavit-Hartenberg (D-H) model based on measured robotic arm structural parameters and integrate hydraulic cylinder dynamics to construct a composite parametric system driven by a Stribeck friction model. By designing high-signal-to-noise ratio displacement excitation signals for hydraulic cylinders and combining Fourier series to construct optimal excitation trajectories that satisfy joint constraints, this method effectively excites the characteristics of each parameter in the minimal parameter set of the dynamic model of the robotic arm. On this basis, a hierarchical progressive parameter identification strategy is proposed: least squares estimation is employed to separately identify and jointly calibrate the dynamic parameters of both the hydraulic cylinder and the robotic arm, yielding Stribeck model curves for each joint. Experimental validation on a robotic arm platform demonstrates residual standard deviations below 0.4 Nm between theoretical and measured joint torques, confirming high-precision dynamic parameter identification for the hydraulic-driven curtain wall installation robotic arm. This significantly contributes to enhancing the intelligence level of curtain wall installation operations.

\end{abstract}

%%%%%%%%%%%%%%%%%%%%%%%%%%%%%%%%%%%%%%%%%%%%%%%%%%%%%%%%%%%%%%%%%%%%%%%%%%%%%%%%
\section{INTRODUCTION}

The dynamic modeling of hydraulic cylinders is essential for hydraulically actuated robotic arms [1]–[3]. In practical engineering applications, the dynamic behavior of hydraulic cylinders is complex and diverse. Friction, in addition to load, is a critical factor that influences performance [4]. Commonly used friction models include the Coulomb-viscous friction model [5], Stribeck friction model [6], Karnopp model [7], and LuGre model [8-9]. The Stribeck model provides greater accuracy than the Coulomb and Coulomb viscous models. Compared with the LuGre model, the Stribeck model offers simpler parameter identification and lower computational complexity. Therefore, we select the Stribeck friction model to describe the friction forces acting on hydraulic cylinders.

Common methods for dynamic modeling include the Newton-Euler iterative method [10], Lagrangian method, Kane's method [11], and screw theory [12]–[13]. Dynamic models incorporate unknown dynamic parameters, with key identification techniques including decomposition measurement, Computer-Aided Design (CAD) measurement [14], and holistic experimental identification [15]–[16]. The CAD measurement technique relies on software calculations, neglecting actual assembly and manufacturing errors. Holistic experimental identification determines the true dynamic parameters of robotic arms by measuring joint displacements and applied torques, and then applying inverse dynamics principles. This method, based on real operational data under actual working conditions, offers high applicability and accuracy for complex robotic arm structures.

\begin{figure}[htb]
      \centering
      \includegraphics[width=\columnwidth]{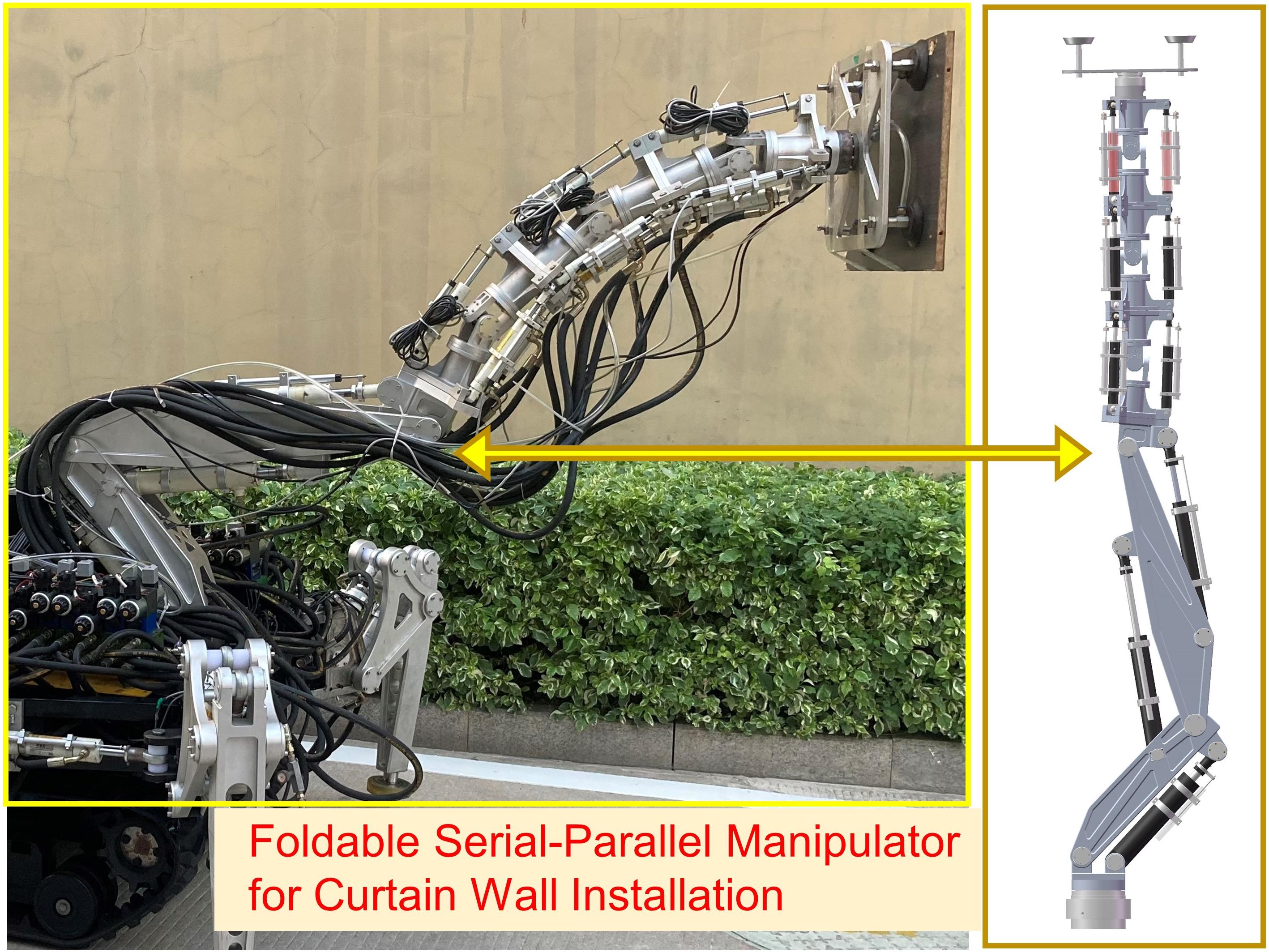}
      \caption{A Glass Curtain Wall Installation Robotic Arm}
      \label{figurelabel}
   \end{figure}
The movements and applied forces of the adjacent joints in robots are coupled, making it challenging to accurately identify each dynamic parameter [17]–[18]. Two common methods can be employed to further determine identifiable dynamic models: a. Lagrangian Method: Construct a linear dynamic model using the Lagrangian approach and transform the regression matrix into an upper triangular form, simplifying the calculation process. b. Parameter Elimination and Reorganization: Remove unidentifiable parameters and reorganize dependent linear parameters. This method has been integrated into libraries such as Symoro and Sympybotics [19], enabling rapid construction of dynamic models and determination of minimal parameter sets by calling these libraries. In recent studies, [20]–[21] compared various methods for dynamic parameter identification and found that the least squares method and adaptive linear neuron (ADALINE) neural networks yielded the best results. [22] proposed a Weighted Least Squares and Random Weighted Particle Swarm Optimization (WLS-RWPSO) algorithm for dynamic parameter identification in collaborative robots. This method achieves a good match between the theoretical and measured torques. The aforementioned studies on dynamic parameter identification are thorough. This paper designs a unique curtain wall installation robotic arm, featuring a serial-parallel manipulator and a large folding arm, all driven by hydraulic cylinders. Therefore, the issue of dynamic parameter identification for complex drive components remains a topic worthy of further discussion and research.
\section{A GLASS CURTAIN WALL INSTALLATION ROBOTIC ARM}
The design of the robotic arm for glass curtain wall installation is illustrated in Figure 1. The foldable robotic arm has a length of 2.3 m. It demonstrates a maximum payload capacity of 40 kg and accommodates panels up to 1 m × 1 m dimensions. It employs remote operation control, enabling convenient multi-angle monitoring during the grasping process. Mounted on a foot-track composite mobile base, the mechanism maintains full operational workspace integrity while ensuring mobility.
\section{ROBOTIC ARM STRUCTURAL PARAMETERS}

\begin{figure}[htb]
      \centering
      \includegraphics[width=\columnwidth]{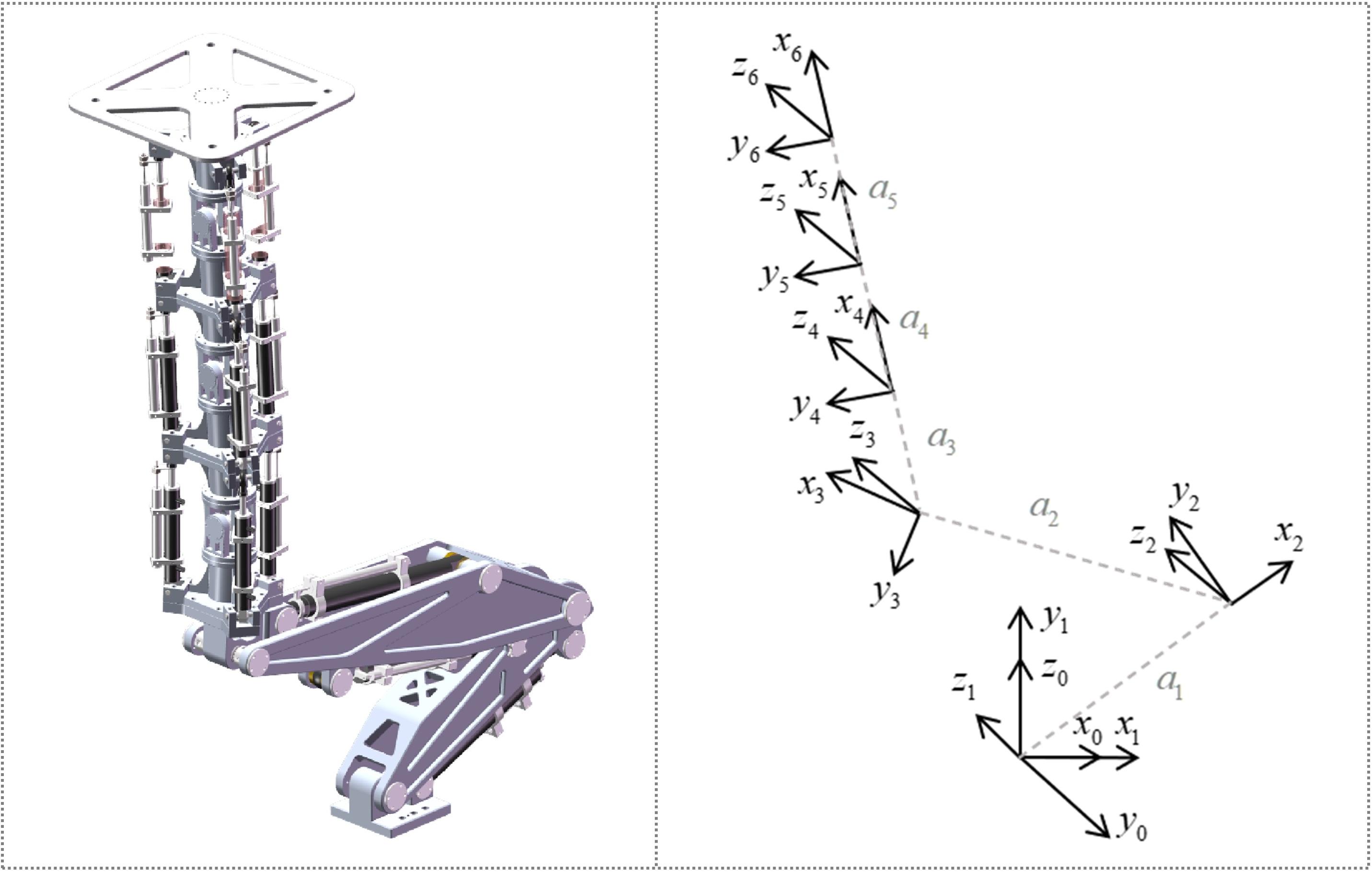}
      \caption{Robotic Arm Model and Link Coordinate Systems}
      \label{figurelabel}
   \end{figure}
As shown in Figure 2, based on the robotic arm's structural parameters, the link coordinate systems can be established. The standard D-H parameters are listed in Table 1.

\begin{table}[ht]
    \centering
    \caption{DH Parameters of the Robotic Arm}
    \resizebox{8.5cm}{!}{%
    \begin{tabular}{ccccc}
        \toprule
             \textbf{Joint $i$} & \textbf{$d_i$ (mm)} & \textbf{$\alpha_i$ ($^\circ$)} & \textbf{$a_i$ (mm)} & \textbf{$\theta_i$ ($^\circ$)} \\
        \midrule
      0 & 0 & 90 & 0 & 0 \\ 
      1 & 0 & 0 & 511 & 30 \\
      2 & 0 & 0 & 842.46 & 130 \\
      3 & 0 & 0 & 245.5 & -60  \\
      4 & 0 & 0 & 300 & 0 \\
      5 & 0 & 0 & 300 & 0  \\
      6 & 0 & 0 & 244.5 & 0 \\ 
        \bottomrule
    \end{tabular}
    }
\end{table}
The joint limits for each joint, including the joint position \(\theta_i\) (rad), maximum angular velocity \( \dot{\theta}_{imax} \) (rad/s), and maximum angular acceleration  \( \ddot{\theta}_{imax} \)  (rad/\( s^2 \)), are specified in Table 2.

\begin{table}[ht]
    \centering
    \caption{Motion Constraints of the Robotic Arm}
    \resizebox{8.5cm}{!}{%
    \begin{tabular}{cccc}
        \toprule
              {Joint $i$} & \textbf{$\theta_i$ (rad)} & \textbf{$\theta'_{i\text{max}}$ (rad/s)} & \textbf{$\theta''_{i\text{max}}$ (rad/s$^2$)} \\
        \midrule
       0 & 0 & 0 & 0 \\
       1 & [-0.0523, 1.0472] & 0.2 & 0.1 \\ 
       2 & [-1.0472, 0.1745] & 0.2 & 0.1 \\
       3 & [-0.8727, 1.5708] & 0.2 & 0.1 \\
       4 & [-0.3491, 0.3491] & 0.5 & 0.8 \\
       5 & [-0.3491, 0.3491] & 0.5 & 0.8 \\
       6 & [-0.3491, 0.3491] & 0.5 & 0.8 \\ 
        \bottomrule
    \end{tabular}
    }
\end{table}

\section{DYNAMIC PARAMETERS IDENTIFICATION FOR ROBOTIC ARM}
We will develop identifiable dynamic models for hydraulic cylinders and robotic arm rigid-body dynamics, and design excitation trajectories to lay the groundwork for parameter identification experiments. The dynamic parameter identification process is illustrated in Figure 3. For complex robotic arm configurations, the computational burden of using disassembly measurement and CAD methods is excessive; hence, we employ a holistic identification approach for dynamic parameter identification. To enhance the accuracy of the overall robotic arm dynamic parameters, we first conduct a separate analysis and parameter identification of the hydraulic cylinder dynamics.
\begin{figure}[htb]
      \centering
      \includegraphics[width=\columnwidth]{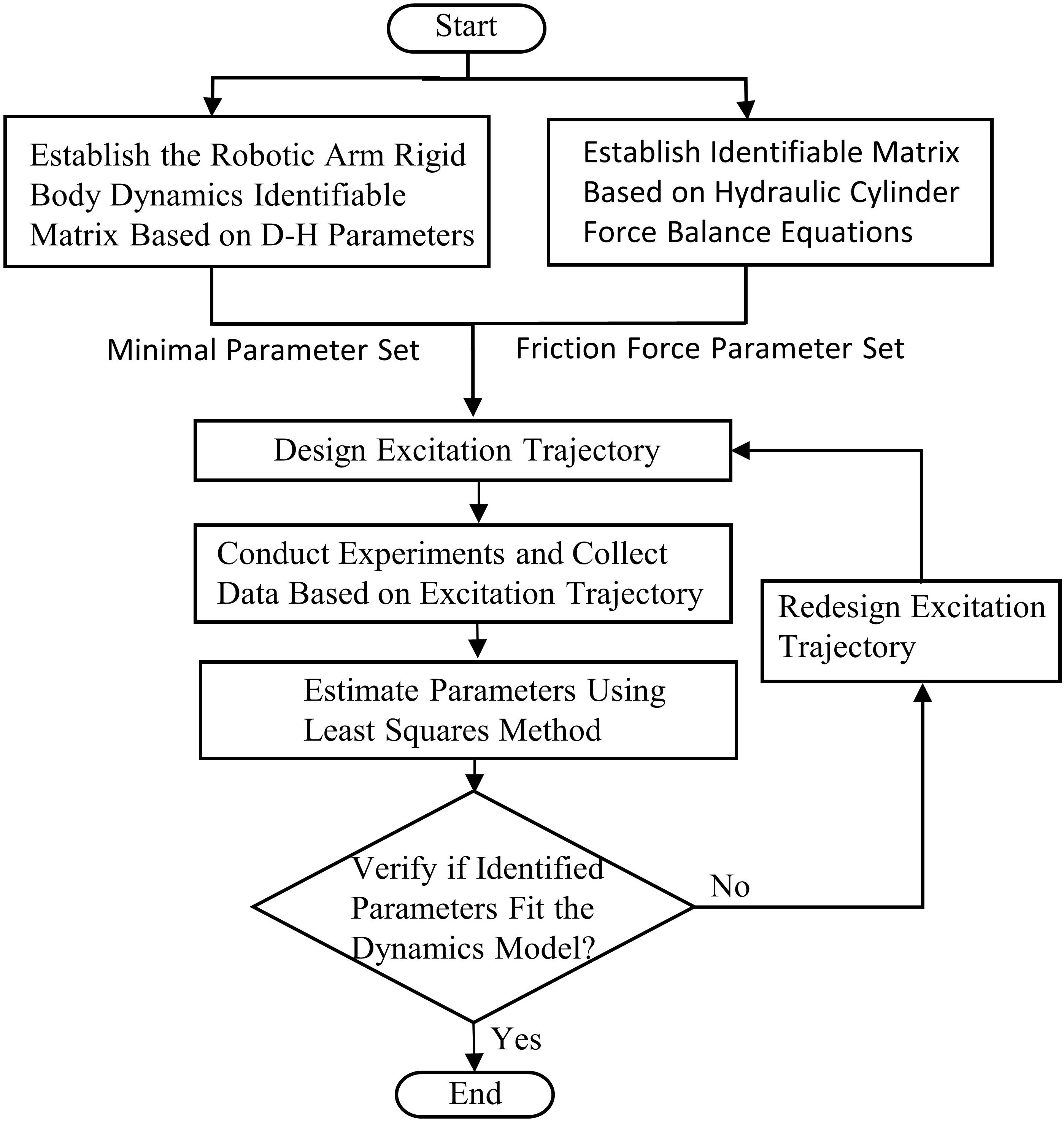}
      \caption{Dynamics Parameter Identification Flow}
      \label{figurelabel}
   \end{figure}

\subsection{Friction Model Identification} The actuation of the robotic arm is via hydraulic cylinders, whose dynamic models are more significantly affected by friction forces compared to motor-driven systems. Therefore, we first perform a separate identification of the friction parameters, specifically focusing on the friction terms in the hydraulic cylinder dynamics. The force balance equation for the hydraulic cylinder system can be expressed as [23] :

\begin{equation}
\label{deqn_ex1}
 m\ddot{x}+F_\mathrm{c}+F_\mathrm{k}+F_\mathrm{f}=p_1A_1-p_2A_2-F 
\end{equation}

In the equation:  \( m \) is the mass of the piston;  \( x \) is the displacement of the piston; \( F_c \) is the viscous force; \( F_k \) is the nonlinear hydraulic elastic force;  \( F_f \) is the nonlinear friction force; \( F \) is the load force acting on the piston; \( p_1 \) and \( p_2 \) are the pressures in the rod side chamber and the cap side chamber, respectively; \( A_1 \) and \( A_2 \) are the effective areas of the rod side chamber and the cap side chamber, respectively.

In a single-rod hydraulic cylinder driven by pressure oil, the piston moves due to the hydraulic elastic effect and must overcome various resistances, including nonlinear dynamic friction forces. The dynamic equation for the hydraulic cylinder can be established as:

\begin{equation}
\label{deqn_ex1}
 m\ddot{x}+c\dot{x}+Kx+F_\mathrm{d}=p_1A_1-p_2A_2-F_\mathrm{L}
\end{equation}

In the equation: \( c \) is the damping coefficient, which is proportional to the velocity;  \( K \) is the stiffness coefficient. We use the Strubeck model to describe the friction force model of the hydraulic cylinder. This model effectively represents the three regions of friction under lubricated conditions [24]: the static friction zone, the transition zone, and the fluid dynamic lubrication zone. The dynamic friction force \( F_d \) can be expressed as:

\begin{equation}
\label{deqn_ex1}
F_d=f_c\operatorname{sgn}(\dot{x})+f_v\dot{x}+\left(f_m-f_c\right)e^{-\left(\frac{\dot{x}}{\nu_s}\right)^s}
\end{equation}

In the equation:  \( f_c \) is the Coulomb friction parameter; \( f_m \) is the maximum static friction force; \( f_v \) is the viscous friction coefficient; \( v_s \) and \( \delta \) are empirical constants.

Linearizing the Strubeck friction model parameters [25] yields:

\begin{equation}
\label{deqn_ex1}
F_\mathrm{d}\left(\dot{x},f_\mathrm{c},f_\mathrm{v},f_\mathrm{s}\right)=f_\mathrm{c}\operatorname{sgn}(\dot{x})+f_\mathrm{v}\dot{x}+f_\mathrm{s}\dot{x}^{1/3}
\end{equation}

Substituting Equation (4) into Equation (1), we obtain:

\begin{equation}
\label{deqn_ex1}
m\ddot{x}+c\dot{x}+Kx+f_\mathrm{c}\operatorname{sgn}(\dot{x})+f_\mathrm{v}\dot{x}+f_\mathrm{s}\dot{x}^{1/3}=p_1A_1-p_2A_2-F
\end{equation}

Set both sides of the hydraulic cylinder system force balance equation to:
\begin{equation}
\label{deqn_ex1}
y(t)=p_1A_1-p_2A_2-F-c\dot{x}
\end{equation}
\begin{equation}
\label{deqn_ex1}
\lambda(t)=\begin{bmatrix}
\dot{x} & x & \mathrm{sgn}(\dot{x}) & \dot{x} & \dot{x}^{1/3}
\end{bmatrix}^\mathrm{T}
\end{equation}
\begin{equation}
\label{deqn_ex1}
\theta=\begin{bmatrix}
m & K & f_\mathrm{c} & f_\mathrm{v} & f_\mathrm{s}
\end{bmatrix}^\mathrm{T}
\end{equation}

In the system, the measurable quantities are \( p_1 \), \( p_2 \), displacement \( x \), the area of the rodless chamber \( A_1 \), and the area of the rod chamber \( A_2 \). The pressures in both chambers vary with piston displacement and can be measured using pressure sensors; the damping coefficient \( c \) is known, so the damping force \( F_c = c \dot{x} \). The piston mass \( m \) is also known, and \( F_L \) can be designed based on the operating conditions. \( K \), \( f_c \), \( f_v \), and \( f_s \) are parameters that need to be estimated.

Thus, Equation (5) can be rearranged into a form suitable for least squares estimation:
\begin{equation}
\label{deqn_ex1}
y(t)=\lambda(t)\alpha+d(t)
\end{equation}
The system output parameters include the piston displacement, pressures in the rodless and rod chambers, and the externally applied rolling force and viscous force. The piston displacement \( x \), velocity \( \dot{x} \), and acceleration \( \ddot{x} \) signals, after being processed by a sign function, form the state regression vector of the system. Discretize the formula:
\begin{equation}
\label{deqn_ex1}
y_k=\lambda_k^\mathrm{T}\alpha+d_k
\end{equation}
In the equation, \( k \) denotes the sampling time. Estimating the parameters of System Equation (5) yields the parameter identification results for the model at time \( k \):
\begin{equation}
\label{deqn_ex1}
K_k=P_{k-1}\lambda_k\left[\lambda_k^\mathrm{T}P_{k-1}\lambda_k-1\right]^{-1}
\end{equation}
\begin{equation}
\label{deqn_ex1}
\hat{\alpha}_k=\hat{\alpha}_{k-1}+K_k\left[y_k-\lambda_k^\mathbf{T}\hat{\alpha}_{k-1}\right]
\end{equation}
\begin{equation}
\label{deqn_ex1}
P_k=\begin{bmatrix}
I-K_k\lambda_k^\mathbf{T}P_{k-1}
\end{bmatrix}
\end{equation}
In the equation, \( \hat{\alpha}_k \) is the parameter estimate at time \( k \). \( K_k \) and \( P_k \) combine predicted states and observation information to update the system state estimate. Based on predefined initial values, by collecting actual observation data and applying recursive algorithms, the parameter estimates are updated and optimized at each iteration step using the parameter estimates from the previous step and the newly acquired data, until the predefined convergence criteria are met, resulting in satisfactory and stable parameter estimates. Using trigonometric functions, a periodically varying hydraulic cylinder displacement signal can be designed, where the restoring force experienced by the piston is proportional to its displacement and always directed towards the equilibrium position.
\begin{figure}[htb]
      \centering
      \includegraphics[width=\columnwidth]{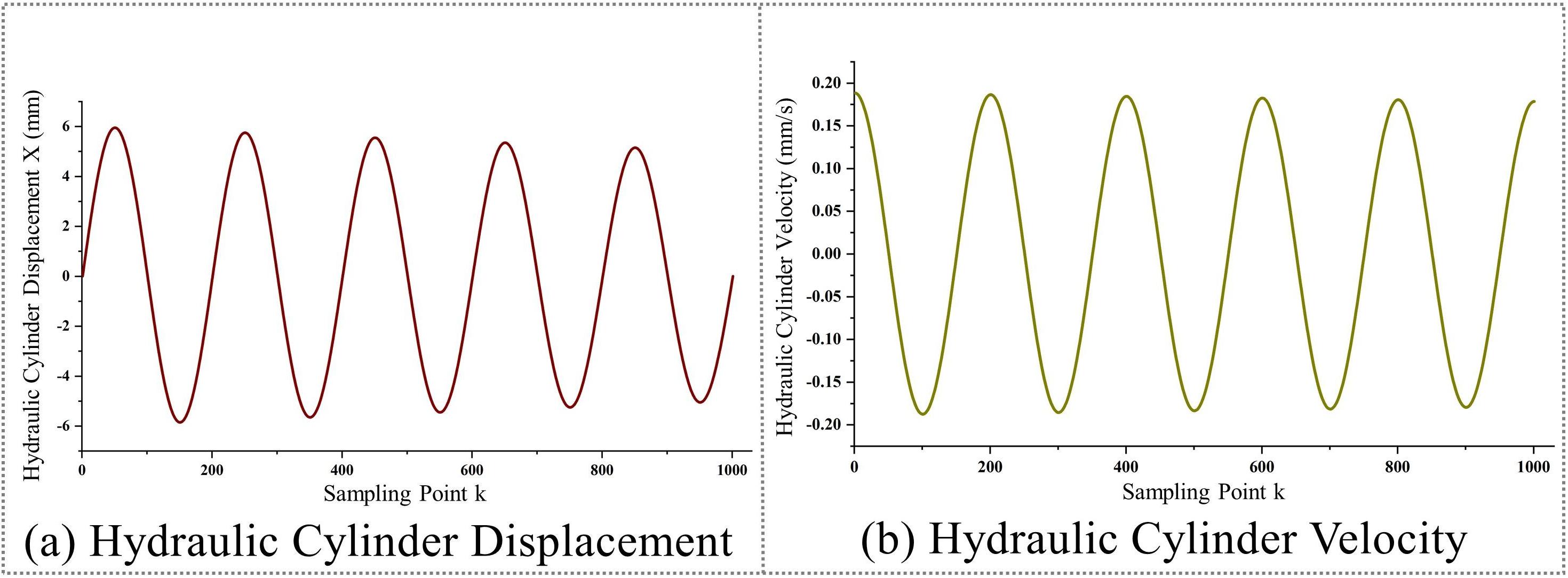}
      \caption{Hydraulic Cylinder Displacement Curve}
      \label{figurelabel}
   \end{figure}
The formulas for hydraulic cylinder displacement and displacement velocity signals are:
\begin{equation}
\label{deqn_ex1}
\begin{aligned}
 & x=6\sin(0.05t) \\
 & \dot{x}=0.3\cos(0.05t)
\end{aligned}
\end{equation}
Positive displacement signals indicate movement towards the rodless chamber, negative signals indicate movement towards the chamber with the rod.
\subsection{Establishing an Identifiable Robotic Arm Model}
The dynamic model established using iterative Newton-Euler equations:
\begin{equation}
\label{deqn_ex1}
\tau = M(q) \ddot{q} + C(q, \dot{q}) \dot{q} + G(q) + \tau_{f}(\dot{q}) \quad
\end{equation}
\[\tau = [\tau_1, \tau_2, \dots, \tau_n]^T\]\(\tau\) is the joint torque vector; \(M(q)\) is the inertia matrix; \(C(q, \dot{q})\) is the Coriolis and centrifugal force matrix; \(G(q)\) is the gravitational force matrix; \(\tau_f(\dot{q})\) is the joint friction torque matrix; \(q = [q_1, q_2, \dots, q_n]^T\) is the joint angle vector, \(\dot{q}\) and \(\ddot{q}\) are the joint velocity and joint acceleration vectors, respectively.
By applying the parallel axis theorem, the inertia tensor can be transformed from one joint coordinate system to the global coordinate system or any other coordinate system parallel to the centroid. This ensures that the new inertia tensor matrix accurately reflects the distribution of the manipulator's mass and rotational inertia in the new coordinate system.
Let \(\{A\}\) be any coordinate system parallel to the center of mass, and \(\{O\}\) be the coordinate system with the origin at the center of mass. Let \(P_o\) be a vector from the coordinate system \(\{A\}\) pointing towards \(\{O\}\). Let \(E\) be the identity matrix. Then, according to the parallel axis theorem, we obtain:
\begin{equation}
\label{deqn_ex1}
\begin{cases}
E^A=E^O+m\left(P_O^TP_O-P_OP_O^T\right) \\
E_i^i=E_i^{Oi}+m\left(P_{Oi}^{iT}P_{Oi}^i-P_{Oi}^iP_{Oi}^{iT}\right) & 
\end{cases}
\end{equation}
After applying Lagrangian transformation and substituting the formulas for inertial forces and driving torques, the dynamic model can be linearized as:
\begin{equation}
\label{deqn_ex1}
\tau_{i}=Y_{i}(q,\dot{q},\ddot{q})X_{i}\quad 
\end{equation}
For any link \( i \), \( Y_i(q, \dot{q}, \ddot{q}) \) denotes the matrix function describing its dynamic characteristics, as the observation matrix. \( X_i \) represents the parameter matrix for the dynamics of the robotic arm.
\begin{equation}
\label{deqn_ex1}
X_{i}=\left[m_{i},I_{xxi},I_{yyi},I_{zzi},I_{xzi},I_{xyi},I_{xzi},r_{xi},r_{yi},r_{zi},f_{ci},f_{vi},f_{si}\right]
\end{equation}
In the equation, \( m_i \) denotes the mass parameter of link \( i \), \((I_{xxi}, I_{yyi}, I_{zzi}, I_{xzi}, I_{xyi}, I_{xzi})\) are the inertia tensor parameters, \((r_{xi}, r_{yi}, r_{zi})\) are the coordinates of the center of mass, and \((f_{ci}, f_{vi}, f_{si})\) are the friction force parameters. By reorganizing \( X_i \), the minimum set of inertia parameters can be obtained. For a six-degree-of-freedom robotic arm, there are 78 parameters to be identified (12 joint friction parameters, 60 inertia parameters, 6 rotational inertia parameters). Therefore, Equation (17) can be rewritten as:
\begin{equation}
\label{deqn_ex1}
\tau=\sum_{i=1}^{78}Y_ix_i
\end{equation}
In this parameter set, if a column in the observation matrix consists entirely of zero elements, it indicates that the corresponding dynamic parameters have no influence on the system's dynamic response and can thus be eliminated. Similarly, if a column is linearly dependent on other columns, it signifies parameter redundancy—where two or more parameters have linearly dependent effects on the system dynamics—and can also be removed. The resultant reduced parameter set can then be expressed as:
\begin{equation}
\label{deqn_ex1}
\tau=Y_1\lambda_1+Y_2\lambda_2+Y_3\lambda_3+\cdots+Y_i\lambda
\end{equation}
The minimal set of inertia parameters is not unique due to the linear dependencies among different parameters in the observation matrix, resulting in multiple solutions with a fixed number of parameters. Parameter reorganization reduces the total number of parameters to be identified without compromising the completeness and accuracy of the dynamic model.

Due to the large computational load, the SymPybotic tool package [26], based on D-H parameters, is utilized to derive the minimal set of inertia parameters and the observation matrix. Since the friction parameters \((f_{ci}, f_{vi}, f_{si})\) have been separately identified, substituting them into the observation matrix yields an observation matrix \(Y(q, \dot{q}, \ddot{q})\) of size 6×18, which can be expressed as:
\begin{equation}
\label{deqn_ex1}
Y(q,\dot{q},\ddot{q})=\begin{bmatrix}Y_{11}&Y_{12}&\cdots&Y_{117}&Y_{118}\\Y_{21}&Y_{22}&\cdots&Y_{217}&Y_{218}\\Y_{31}&Y_{32}&\cdots&Y_{317}&Y_{318}\\Y_{41}&Y_{42}&\cdots&Y_{417}&Y_{418}\\Y_{51}&Y_{52}&\cdots&Y_{517}&Y_{518}\\Y_{61}&Y_{62}&\cdots&Y_{617}&Y_{618}\end{bmatrix}
\end{equation}
Observing the matrix above, it can be seen that the minimal set of inertia parameters should consist of an 18×1 vector, comprising 18 independent parameters. Compared to the original 78 parameters requiring identification, this reduces the number by 60, significantly lowering the complexity and difficulty of the inertia parameter identification process. This simplification makes system modeling and analysis more straightforward and efficient.
The identification equations can be organized as:
\begin{equation}
\label{deqn_ex1}
\Gamma(\tau)=H(q,\dot{q},\ddot{q})X+\rho
\end{equation}
In the equation, \(\Gamma(\tau)\) denotes the measured force vector; \(H(q, \dot{q}, \ddot{q})\) represents the observation matrix for the identification equation; \(\rho\) is the error vector, used to represent the difference between the actual measured values of joint angles, positions, velocities, or accelerations and their target values. The measured force vector \(\Gamma(\tau)\) and the observation matrix \(H(q, \dot{q}, \ddot{q})\) can be expressed as:
\begin{equation}
\label{deqn_ex1}
\begin{aligned}
 & \Gamma(\tau)=
\begin{bmatrix}
\Gamma^1 \\
\Gamma^2 \\
\vdots \\
\Gamma^6
\end{bmatrix},\Gamma_i=
\begin{bmatrix}
\tau_i\left(t_1\right) \\
\tau_i\left(t_2\right) \\
\vdots \\
\tau_i\left(t_k\right)
\end{bmatrix} \\
 & H(q,\dot{q},\ddot{q})=
\begin{bmatrix}
H_1 \\
H_2 \\
\vdots \\
H_6
\end{bmatrix},H_i=
\begin{bmatrix}
H\left(q\left(t_1\right),\dot{q}\left(t_1\right),\ddot{q}\left(t_1\right)\right) \\
H\left(q\left(t_2\right),\dot{q}\left(t_2\right),\ddot{q}\left(t_2\right)\right) \\
\vdots \\
H\left(q\left(t_k\right),\dot{q}\left(t_k\right),\ddot{q}\left(t_k\right)\right)
\end{bmatrix}
\end{aligned}
\end{equation}
In the equation, \( i = 1, 2, \ldots, 6 \) denotes the joint index of the robotic arm; \( k \) is the total number of sampled points; \( \Gamma_i \) represents the set of measured torque values \( \tau(t) \) corresponding to joint \( i \); where \( \tau_i(t) \) is the \( i \)-th element of the vector \( \tau(t) \) at time \( t \); \( H_i \) is the subset of the observation matrix \( H(q(t), \dot{q}(t), \ddot{q}(t)) \) corresponding to joint \( i \).

The optimal set of parameters \(\hat{\beta}\) is identified using the least squares method, minimizing the squared difference between the measured torque and the theoretical torque, i.e., \(\|\rho\|^2\), to find the best estimate of the model parameters. Equation (24) can be expressed in the form of a least squares estimate as:
\begin{equation}
\label{deqn_ex1}
\hat{\beta}=\left(H^{T}H\right)^{-1}H^{T}\Gamma 
\end{equation}
\subsection{Design of Excitation Trajectories for Robotic Arm Identification}
In the parameter identification process for robotic arm dynamic models, designing appropriate excitation trajectories is crucial. To ensure accurate parameter identification, excitation trajectories must adequately stimulate all dynamic characteristics of the robotic arm, such as performance at various speeds and accelerations. Additionally, the designed trajectories should help distinguish between different parameters in the model, reducing parameter correlations and enhancing identification accuracy and reliability.

We construct the excitation trajectories for the robotic arm based on Fourier series, with the model represented as:
\begin{equation}
\label{deqn_ex1}
\begin{aligned}
 & q_{i}(t)=\sum_{l=1}^{n_H}\left(\frac{a_{il}}{\omega_fl}\sin\left(\omega_flt\right)-\frac{b_{il}}{\omega_fl}\cos\left(\omega_flt\right)\right)+q_{i,0} \\
 & \dot{q}_i(t)=\sum_{l=1}^{n_H}\left(a_{il}\cos\left(\omega_flt\right)+b_{il}\sin\left(\omega_flt\right)\right) \\
 & \ddot{q}_{i}(t)=\sum_{l=1}^{n_H}\left(-a_{il}w_fl\sin\left(\omega_flt\right)+b_{il}w_fl\cos\left(\omega_flt\right)\right)
\end{aligned}
\end{equation}
In the equation, \( t \) denotes time; \( q_i(t) \), \( \dot{q}_i(t) \), and \( \ddot{q}_i(t) \) represent the joint angle, angular velocity, and angular acceleration of joint \( i \) at time \( t \), respectively; \( w_f \) is the angular frequency of the Fourier series; \( l = 1, 2, \ldots, n_H \) is the harmonic level number of the trajectory; \( a_l^i \) and \( b_l^i \) are the amplitudes of the trigonometric functions in the trajectory; \( q_{i,0} \) is the initial offset of the joint angle. Under the constraint conditions, a reasonable setting of \( q_{i,0} \) allows for a larger angular frequency in the Fourier series, enabling the excitation trajectory to better stimulate the dynamic performance of the robotic arm and improve the signal-to-noise ratio.

When designing excitation trajectories, it is essential to ensure compliance with joint limits and achieve smooth transitions. This avoids noticeable discontinuities or jitter in the motion curves and reduces shock loads. In summary:
\begin{equation}
\label{deqn_ex1}
\begin{gathered}
\left|q_i(t)\right|\leq{q}_{i,\max} \\
\left|\dot{q}_i(t)\right|\leq\dot{q}_{i,\max} \\
\left|\ddot{q}_i(t)\right|\leq\ddot{q}_{i,\max} \\
q_i\left(t_0\right)=q_i\left(t_f\right)=0 \\
\dot{q}_i\left(t_0\right)=\dot{q}_i\left(t_f\right)=0 \\
\ddot{q}_i\left(t_0\right)=\ddot{q}_i\left(t_f\right)=0
\end{gathered}
\end{equation}
In the equation, \( t_0 \) denotes the start time of the trajectory, and \( t_f \) denotes the end time of the trajectory. The objective function of the excitation trajectory is to minimize the condition number of the identification matrix. Based on the observation matrix, the minimum condition number of the matrix can be determined, where the constraint conditions can be expressed as:
\begin{equation}
\label{deqn_ex1}
\begin{aligned}
 & \left|q_i(t)\right|=\left|\sum_{l=1}^N\left(\frac{a_{il}}{\omega_fl}\sin\left(\omega_flt\right)-\frac{b_{il}}{\omega_fl}\cos\left(\omega_flt\right)\right)+q_{i0}\right| \\
 & \leq\sum_{l=1}^N\frac{1}{\omega_fl}\sqrt{{a_{il}}^2+{b_{il}}^2}+\left|q_{i0}\right|\leq{q}_{i,\max}
\end{aligned}
\end{equation}
In the experiment, the excitation trajectory for the robotic arm is set with an angular frequency of \( \omega_f = 0.1\pi \, \text{Hz} \) and a period of \( T_f = 20 \, \text{s} \). For the robotic arm, the angular frequency is set to \( \omega_f = 0.2\pi \, \text{Hz} \) with a period of \( T_f = 10 \, \text{s} \). \( n_H \) denotes the level number in the Fourier series, which is set to \( n_H = 3 \) in the experiment. This means that each joint has 7 parameters to be determined, resulting in a total of 42 parameters to be solved.

Based on the obtained parameters, continuous curves for the position, velocity, and acceleration of each joint over time can be constructed, as shown in Figure 5.
\begin{figure}[htb]
      \centering
      \includegraphics[width=\columnwidth]{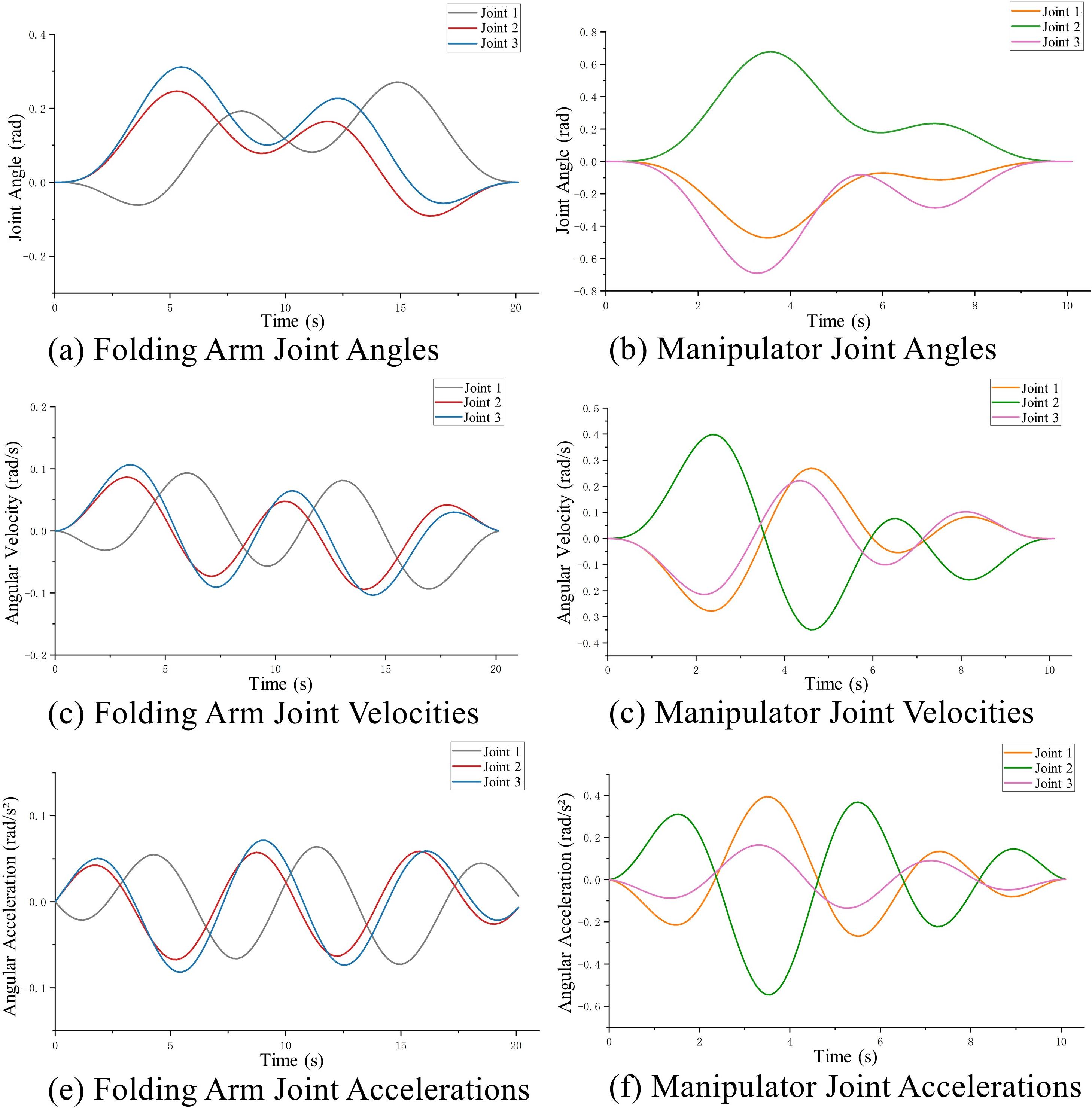}
      \caption{Excitation Parameter Trajectories of Each Joint}
      \label{figurelabel}
   \end{figure}
The parameters \( a_il \), \( b_il \), and \( q_{i,0} \) for the Fourier series are specified in Table 3.

\begin{table}[h]
\centering
\caption{Setting of Fourier Leaf Numbers \( a_{il} \), \( b_{il} \), and \( q_{i,0} \) for Each Joint}
\label{tab:joint-variables}
\begin{tabular}{@{}ccccccc@{}}
\toprule
Joint Variable & Joint 1 & Joint 2 & Joint 3 & Joint 4 & Joint 5 & Joint 6 \\ \midrule
$a_{i1}$      & -8.996  & 8.615   & 5.486   & -5.669  & 4.879   & -4.169  \\
$b_{i1}$      & 8.600   & 6.350   & 5.941   & -4.520  & 7.538   & -4.151  \\
$a_{i2}$      & -8.464  & 6.524   & 5.265   & 7.319   & -8.006  & 4.731   \\
$b_{i2}$      & 8.106   & 3.476   & 5.936   & -4.702  & 6.429   & -8.877  \\
$a_{i3}$      & 17.460  & -15.138 & -10.751 & -1.650  & 3.127   & -0.562  \\
$b_{i3}$      & -8.270  & -4.434  & -5.938  & 4.642   & -6.798  & 7.302   \\
$q_{i,0}$     & 31.500  & 21.038  & 22.058  & -8.473  & 13.507  & -9.797  \\ \bottomrule
\end{tabular}
\end{table}

Observations from Figures 5 and Table 3 indicate:
(1) Excitation signal amplitudes are sufficient for noticeable dynamic responses while staying within safe velocity and acceleration limits. (2) The duration of the excitation trajectory signals should be long enough for the robotic arm's dynamic system to fully respond to the excitation, facilitating accurate extraction of dynamic characteristics from the response data.

\section{DYNAMICS PARAMETER IDENTIFICATION EXPERIMENT FOR ROBOTIC ARM}
We identified the hydraulic cylinder dynamics parameters and robotic arm rigid-body dynamics parameters using the least squares method based on our established parameter identification framework. Next, we will validate the theoretical torques calculated from the identified dynamics parameters against the measured torques.
\subsection{Robotic Arm Experimental Platform}
The main components of the robotic arm include hydraulic cylinders, multi-channel proportional valve modules, hydraulic pumps, hydraulic motors, sensors, and others, as shown in Figure 6.
\begin{figure}[htb]
      \centering
      \includegraphics[width=\columnwidth]{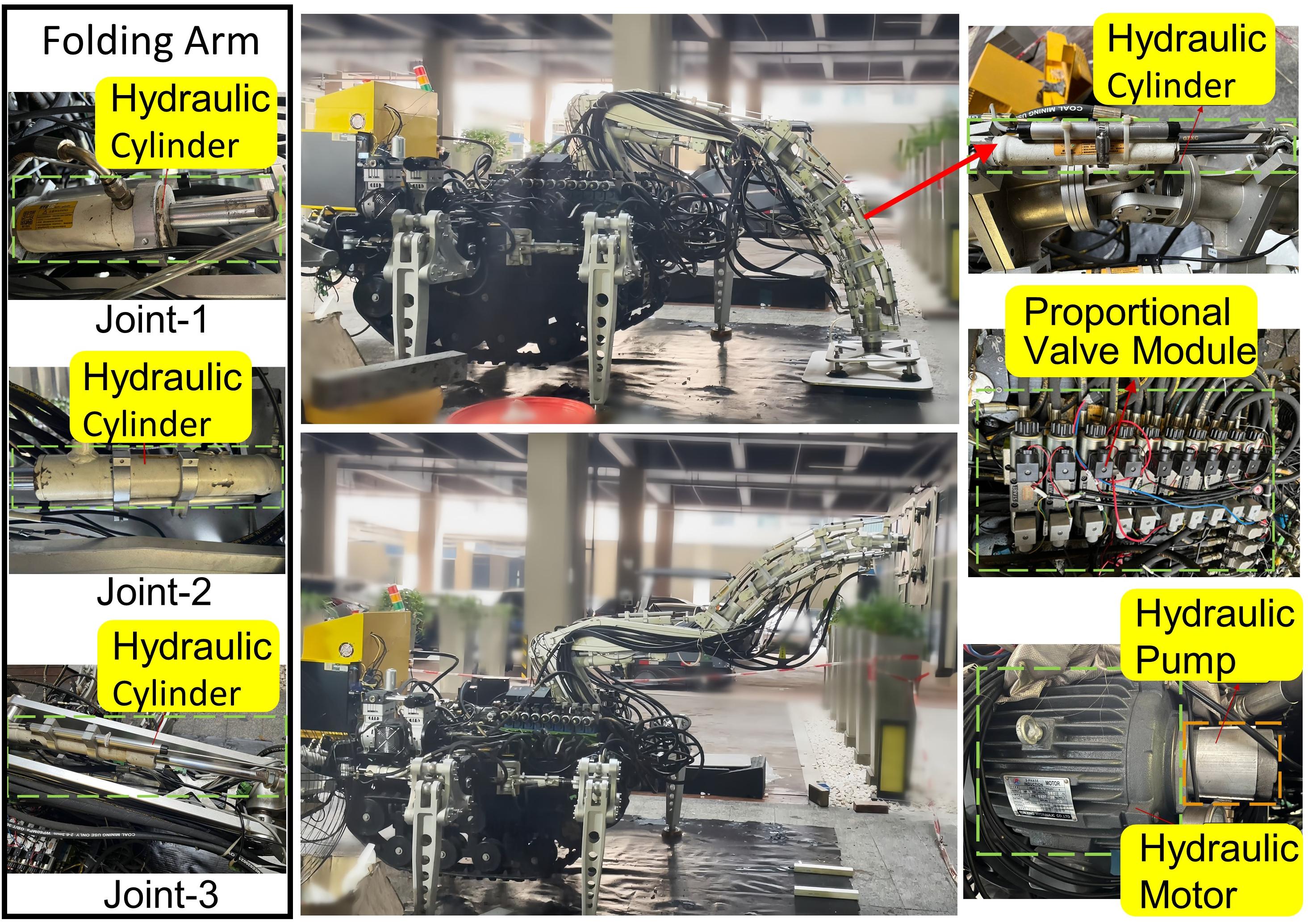}
      \caption{Experimental Platform and Key Components}
      \label{figurelabel}
   \end{figure}
\subsection{Hydraulic Cylinder Dynamics Parameter Identification Experiment}
First, we conducted a parameter identification experiment for the hydraulic cylinder dynamics model. During the experiment, signals from displacement sensors and pressure sensors installed on the robotic arm joints were collected at a sampling frequency of 50 Hz. Throughout the experiment, all joints were unloaded, resulting in zero external forces; the viscosity coefficients were set to \( c = 2 \); and the piston masses are specified in Table 4.
\begin{table}[h]
\centering
\caption{Piston Mass Settings}
\label{tab:piston-mass-settings}
\begin{tabular}{@{}ccccccc@{}}
\toprule
Joint $i$ & 1 & 2 & 3 & 4 & 5 & 6 \\ \midrule
Piston mass $m$ (kg) & 4.595 & 4.914 & 2.324 & 1.937 & 1.729 & 1.705 \\ \bottomrule
\end{tabular}
\end{table}
Six sets of Stribeck model parameter curves can be plotted based on the estimated parameters, as shown in Figure 7:
\begin{figure}[htb]
      \centering
      \includegraphics[width=\columnwidth]{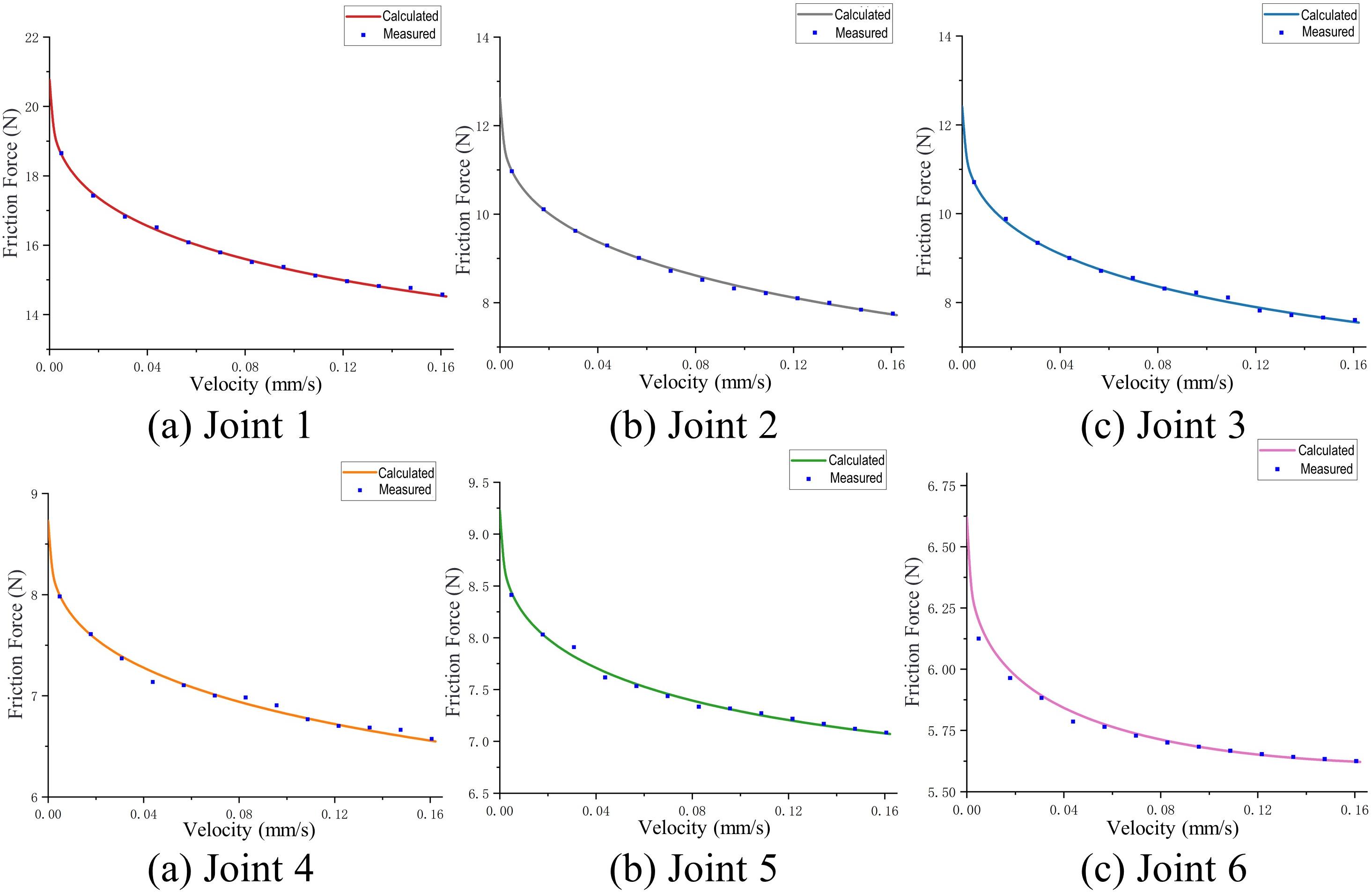}
      \caption{Stirbeck Model Curves of Each Joint}
      \label{figurelabel}
   \end{figure}
Comparing the experimental data with the fitted values from the constructed model reveals a high degree of conformity. Joints four, five, and six show significant deviations in Stribeck model curves despite identical configurations, primarily due to manufacturing tolerances and component fitting variations. Subsequent integration with rigid body dynamics models to further validate its accuracy in predicting joint torques. Detailed Stribeck model curve parameters are provided in Table 5.
\begin{table}[h]
\centering
\caption{Strubeck Friction Model Parameter Sets for Each Joint's Hydraulic Cylinder}
\label{tab:strubeck-friction-parameters}
\begin{tabular}{@{}cccc@{}}
\toprule
Joint $i$ & $f_{c,i}$ (N$\cdot$m) & $f_{v,i}$ (N$\cdot$m) & $f_{s,i}$ (N$\cdot$m) \\ \midrule
1         & 20.77                 & 7.83                  & -15.15               \\
2         & 12.62                 & 4.51                  & -11.53               \\
3         & 12.41                 & 7.03                  & -11.99               \\
4         & 8.73                  & 2.23                  & -5.18                \\
5         & 9.23                  & 4.41                  & -5.60                \\
6         & 6.62                  & 3.99                  & -3.01                \\ \bottomrule
\end{tabular}
\end{table}
\subsection{Robotic Arm Dynamics Parameter Identification Experiment}
The relevant dynamic parameters, as shown in Table 6, were estimated using the least squares method.
\begin{table}[h]
\centering
\caption{Minimal Inertia Parameter Set $\hat{\beta}$ for Robotic Arm}
\label{tab:inertia-parameters}
\begin{tabular}{@{}ccc@{}}
\toprule
Number & Identification Parameters & Identification Values \\ \midrule
1      & $I_{1zz} + 0.261(m_1 + m_2 + m_3 + m_4 + m_5 + m_6)$ & $20.673 \, (kg \cdot m^2)$ \\
2      & $r_{x1} + 0.511(m_1 + m_2 + m_3 + m_4 + m_5 + m_6)$ & $26.448 \, (kg \cdot m^2)$ \\
3      & $r_{z1}$ & $0.418 \, (kg \cdot m^2)$ \\
4      & $I_{2yy} - 0.710(m_2 + m_3 + m_4 + m_5 + m_6)$ & $-26.148 \, (kg \cdot m^2)$ \\
5      & $r_{x2} + 0.842(m_2 + m_3 + m_4 + m_5 + m_6)$ & $31.967 \, (kg \cdot m^2)$ \\
6      & $r_{z2}$ & $0.759 \, (kg \cdot m^2)$ \\
7      & $I_{3yy} - 0.060(m_3 + m_4 + m_5 + m_6)$ & $-0.948 \, (kg \cdot m^2)$ \\
8      & $r_{x3} + 0.245(m_3 + m_4 + m_5 + m_6)$ & $6.120 \, (kg \cdot m^2)$ \\
9      & $r_{z3}$ & $0.457 \, (kg \cdot m^2)$ \\
10     & $I_{4yy} - 0.009(m_4 + m_5 + m_6)$ & $-0.109 \, (kg \cdot m^2)$ \\
11     & $r_{x4} + 0.300(m_4 + m_5 + m_6)$ & $4.878 \, (kg \cdot m^2)$ \\
12     & $r_{z4}$ & $0.084 \, (kg \cdot m^2)$ \\
13     & $I_{5yy} - 0.009(m_5 + m_6)$ & $-0.061 \, (kg \cdot m^2)$ \\
14     & $r_{x5} + 0.300(m_5 + m_6)$ & $3.131 \, (kg \cdot m^2)$ \\
15     & $r_{z5}$ & $0.078 \, (kg \cdot m^2)$ \\
16     & $I_{6yy} - 0.060m_6$ & $-0.279 \, (kg \cdot m^2)$ \\
17     & $r_{x6} + 0.245m_6$ & $1.290 \, (kg \cdot m^2)$ \\
18     & $r_{z6}$ & $0.079 \, (kg \cdot m^2)$ \\ \bottomrule
\end{tabular}
\end{table}
After obtaining the dynamic parameters of the robotic arm, the actual position, velocity, and acceleration data were input into the observation matrix to calculate the theoretical torque values for each joint. These theoretical torques were then compared with the measured torques. The results are shown in Figure 8:
\begin{figure}[htb]
      \centering
      \includegraphics[width=\columnwidth]{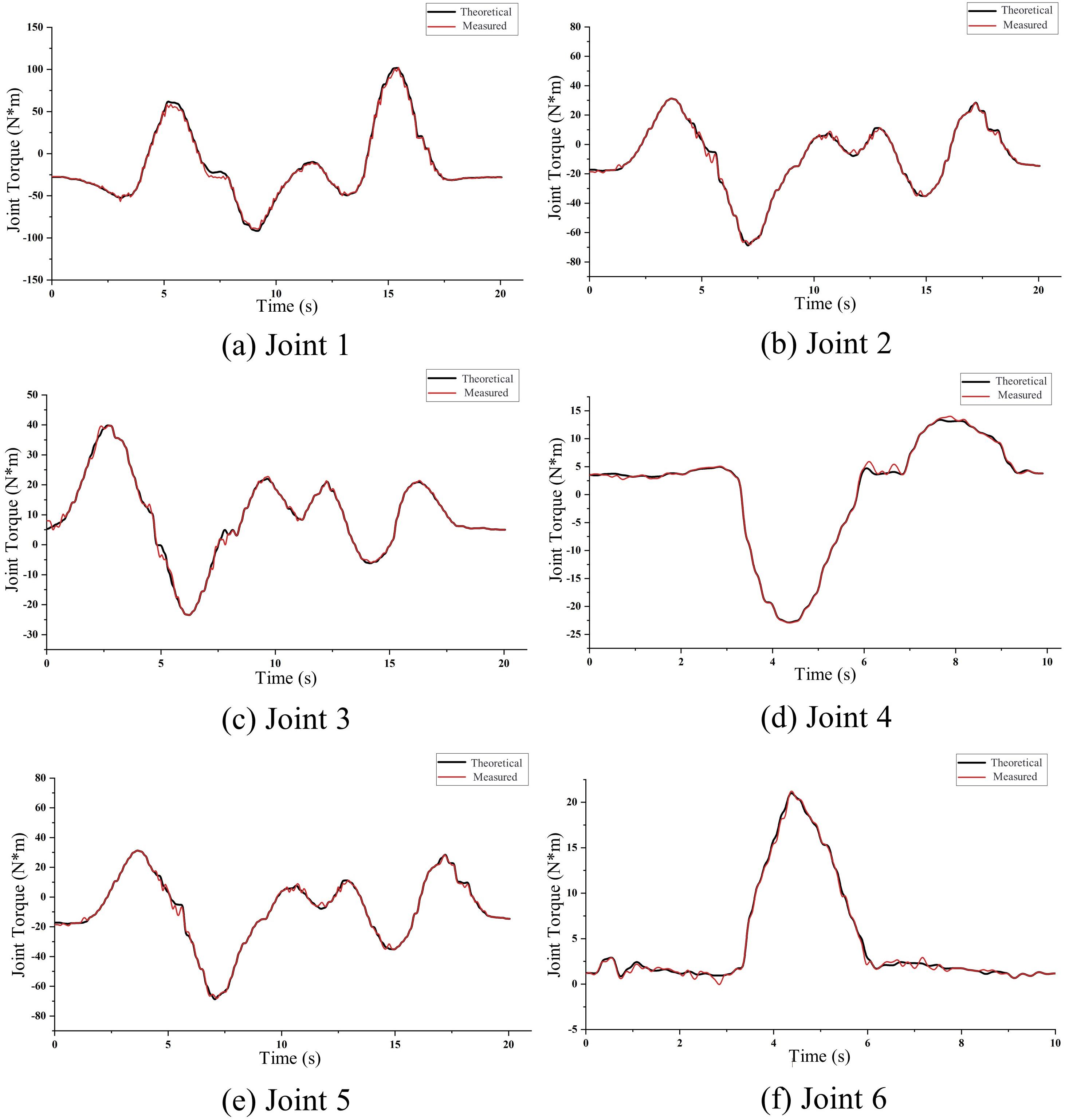}
      \caption{Comparison of Theoretical and Measured Torques for Each Joint}
      \label{figurelabel}
   \end{figure}
To evaluate the discrepancy between theoretical and actual torques, the residual standard deviation (RSD) is used as the error metric. RSD quantifies the dispersion between model predictions and observed values; a smaller RSD indicates higher consistency between the model's predictions and actual observations. The formula for calculating RSD is:
\begin{equation}
\label{deqn_ex1}
\xi_{RSD}=\sqrt{\frac{\sum_{k=1}^K\left(\tau_{ik}-\tau_{ik}^{\prime}\right)^2}{\sum_{k=1}^K\tau_{ik}^2}}
\end{equation}
In the equation: \( k \) is the sampling point; \( K \) is the total number of sampling points; \( \tau_{ik} \) is the estimated torque for joint \( i \) at sampling point \( k \); \( \tau'_{ik} \) is the measured torque for joint \( i \) at sampling point \( k \). The residual standard deviations of the torques for each joint are shown in Table 7.
\begin{table}[h]
\centering
\caption{Residual Standard Deviation of Joint Torques for
 Robotic Arm}
\label{tab
:residual-standard-deviation}
\begin{tabular}{@{}ccccccc@{}}
\toprule
Joint & Joint 1 & Joint 2 & Joint 3 & Joint 4 & Joint 5 & Joint 6
 \\ \midrule
$\xi_{RSD}(\mathrm{Nm})$  & 0.226 & 0.289 & 0.378 & 0.1929 & 0.1787 & 0.3356
 \\ \bottomrule
\end{tabular}
\end{table}
The residual standard deviation between the theoretical torques derived from the dynamic model and the actual torques measured by sensors is less than 0.4 Nm, validating the accuracy of the dynamic parameters.

\section{CONCLUSIONS}

This study proposes a hydraulic-driven robotic arm for curtain wall installation and its dynamic parameter identification framework. The Denavit-Hartenberg (D-H) model is derived from the measured structural dimensions. A hydraulic cylinder dynamics model integrating the Stribeck friction model is formulated, and hydraulic cylinder displacement signals are generated using least squares fitting. Subsequently, an identifiable robotic arm dynamic model was then constructed, and Fourier series-based excitation trajectories are designed that satisfy joint constraint conditions, effectively exciting the characteristics of each parameter in the minimal parameter set of the dynamic model. Least squares estimation was employed to separately identify and jointly calibrate the dynamic parameters of both the hydraulic cylinder and the robotic arm, yielding Stribeck model curves for each joint. Experimental validation in real-world scenarios demonstrated that the residual standard deviations between theoretical and measured joint torques are below 0.4 Nm, confirming the high precision of the dynamic parameter identification for the hydraulic-driven curtain wall installation robotic arm. This work contributes significantly to enhancing the automation and intelligence of curtain wall installation.

\addtolength{\textheight}{-8cm}   % This command serves to balance the column lengths
                                  % on the last page of the document manually. It shortens
                                  % the textheight of the last page by a suitable amount.
                                  % This command does not take effect until the next page
                                  % so it should come on the page before the last. Make
                                  % sure that you do not shorten the textheight too much.

%%%%%%%%%%%%%%%%%%%%%%%%%%%%%%%%%%%%%%%%%%%%%%%%%%%%%%%%%%%%%%%%%%%%%%%%%%%%%%%%

%%%%%%%%%%%%%%%%%%%%%%%%%%%%%%%%%%%%%%%%%%%%%%%%%%%%%%%%%%%%%%%%%%%%%%%%%%%%%%%%

%%%%%%%%%%%%%%%%%%%%%%%%%%%%%%%%%%%%%%%%%%%%%%%%%%%%%%%%%%%%%%%%%%%%%%%%%%%%%%%%

\section*{ACKNOWLEDGMENT}

This work was supported in part by the National Key R\&D Program of China (No.2023YFB4705002), in part by the National Natural Science Foundation of China(U20A20283), in part by the Guangdong Provincial Key Laboratory of Construction Robotics and Intelligent Construction (2022KSYS 013), in part by the CAS Science and Technology Service Network Plan (STS) - Dongguan Special Project (Grant No. 20211600200062), in part by the Science and Technology Cooperation Project of Chinese Academy of Sciences in Hubei Province Construction 2023.

\end{document}